\documentclass[preprint,review,12pt]{elsarticle}

\usepackage{amssymb}
\usepackage{epsfig}
\usepackage{algorithm}
\usepackage{algorithmic}
\usepackage{subfigure}
\usepackage{calc}
\usepackage{amssymb}
\usepackage{amstext}
\usepackage{amsmath}
\usepackage{multicol}

\journal{Journal of Computational Science}

\begin{document}

\begin{frontmatter}

\title{A two-phase decision support framework for the automatic screening of digital fundus images}

\author[ab,inf]{B\'alint Antal\corref{cor1}}
\ead{antal.balint@inf.unideb.hu}
\author[inf]{Andr\'as Hajdu}
\author[med]{Zsuzsanna Maros-Szab\'o}
\author[med]{Zsolt T\"or\"ok}
\author[med]{Adrienne Csutak}
\author[moo]{T\"unde Pet\H o}
\address[inf]{University of Debrecen, Faculty of Informatics, 4010 Debrecen, POB 12, Hungary,
  }
\address[med]{University of Debrecen, Medical and Health Science Centre, 4032 Debrecen, Nagyerdei Krt. 98}
\address[moo]{Moorfields Eye Hospital, London, United Kingdom}
\cortext[cor1]{Corresponding author}
\begin{abstract}
In this paper we give a brief review on the present status of automated detection systems describe  for the screening of diabetic retinopathy. We further detail an enhanced detection procedure that consists of two steps. First, a pre-screening algorithm is considered to classify the input digital fundus images based on the severity of abnormalities. If an image is found to be seriously abnormal, it will not be analysed further with robust lesion detector algorithms. As a further improvement, we introduce a novel feature extraction approach based on clinical observations. The second step of the proposed method detects regions of interest with possible lesions on the images that previously passed the pre-screening step. These regions will serve as input to the specific lesion detectors for detailed analysis. This procedure can increase the computational performance of a screening system. Experimental results show that both two steps of the proposed approach are capable to efficiently exclude a large amount of data from further processing, thus, to decrease the computational burden of the automatic screening system.

\end{abstract}

\begin{keyword}
Biomedical image processing, Medical decision-making, Medical expert systems

\end{keyword}

\end{frontmatter}

\section{\uppercase{Introduction}}
\label{sec:intro}

Retinal fundus photography is widely used at the diagnosis and at regular controls of the consequent treatment of various eye diseases, such as diabetic retinopathy (DR), age related macular degeneration (AMD) and glaucoma. DR is one of the most frequent causes of visual impairment in developed countries and is the leading cause of new cases of blindness among those in the working age \cite{vidr}. In 1997 an estimated 124 million people had diabetes worldwide. This is expected to nearly double by 2010. At any point in time, approximately 40\% of persons with diabetes have diabetic retinopathy, of which an estimated 5\% have the sight-threatening form of this disease. Altogether every day nearly 75 people go blind from DR even though treatment is available \cite{causes}.

Timely detection, organized and practised screening programs are the mainstay of identifying patients at risk for developing any symptoms of DR. Several countries elaborated nation or region wide programs to fulfil this goal. In the United States \cite{vanderbilt} \cite{eyepacs}, in the United Kingdom \cite{nhs} and in The Netherlands \cite{eyecheck} there are digital photography acquisition and reader centre sites already available in daily routine. Color digital retinal images are captured at serviceable sites even outside of health care settings and data will then be transferred to central locations where they are double read and evaluated by specially trained graders. Further health provision of the patient depends upon the result of the grading.

Automated grading of DR based on the detection of the characteristic lesions would safely reduce the burden of manual grading in screening programs. Promising results on higher sensitivity compared with manual graders are already reported in \cite{Philip} for patients having referable diabetic retinopathy. Although the overall specificity of automated grading was lower than the manual analysis, remarkable financial savings could be achieved by reducing the grading workload. Screening programs can be organized to reduce the risk of the disease within the population. Though with the automated screening systems we have to make a compromise between sensitivity and specificity, an alternate approach with high performance is currently not available to provide mass screening. So for a fair competition, automated screening systems should be compared by human experts routinely involved. In this sense, manual grading is imperfect, since graders required to be highly specific missed more than 5\% of the cases of referable diabetic retinopathy in a study \cite{Philip}.

With an automated decision support of the grading process wider access could be provided to the service and improvements could be realized at personal and community DR care level. Graders in DR reading centres are taught to recognize patterns which represent lesions like microaneurysm, dot and blot haemorrhages, lipid exudates and cotton wool spots. With the implementation of computer aided pattern recognition algorithms, the detection of the above mentioned lesions would be theoretically possible. In the past, much effort has been made by different research groups to develop mature algorithms, with sensitivity and specificity is close to that of humans \cite{Winder}. The performance of the algorithms is approaching their limit, though their use is not recommended for clinical practice \cite{system} yet. However, the developments in the field High-performance computing can advance the spread of such methods \cite{Chorley} \cite{Bader}. 
                   
The process of analysing fundus images may be performed by a series of steps. For each step, a number of different approaches are available. It is very hard to determine which are the best algorithms to employ at each step since there is no gold-standard or consensus even in the detection of the regions of interest. In a literature review performed by Winder et al. \cite{Winder} the following pattern recognition steps were identified creating the detection algorithmic sequence: preprocessing, localization and segmentation of the optic disc, segmentation of the retinal vasculature, localization of the macula and fovea, detection of retinopathy. Our interest is to promote the automated decision support framework by means of inserting a preliminary pre-filtering phase before the detailed analysis. Such algorithms usually aim to detect low quality images. For example, in \cite{niemeijer}, a such approach is proposed. In this paper, we propose two new stjpg which can be considered in this phase: pre-screening and pre-filtering.

\begin{figure}[htb]
\centering
\subfigure[]{\includegraphics[keepaspectratio,width=5cm]{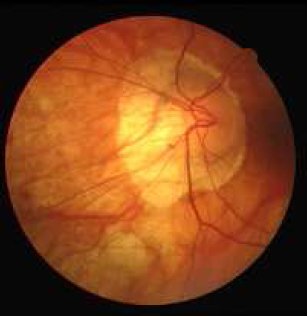}}
\quad\quad
\subfigure[]{\includegraphics[keepaspectratio,width=5cm]{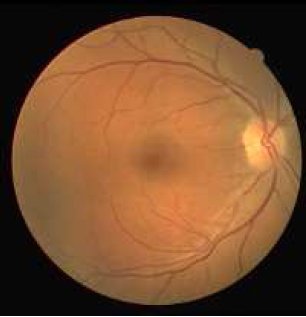}}
\caption{Samples from the image set (both taken from the DRIVE database \cite{drive}); (a) abnormal fundus, (b) not abnormal fundus, the proper grading needs further analysis.}
\label{fig:lab}
\end{figure}

During pre-screening, we classify the images as severely diseased (highly abnormal) or to be forwarded for further processing. The aim of this step is twofold. On the one hand, we minimize the risk that an abnormal image passes the screening without a warning, since it is immediately spotted by the automatic system before detailed analysis. On the other hand, we save computational time, since only the not obviously abnormal fundus images are analysed in details. Figure \ref{fig:lab} gives an impression about these two classes. At the analysis of fundus images, machine learning algorithms are often applied to classification based on feature vectors consisting of intensity values of the image in other fields, see e.g. \cite{hiv} for HIV or \cite{nyul} for glaucoma detection. Thus, we consider implementing these approaches for DR screening, as well. We also improved the techniques with feature extraction based on the inhomogeneity characteristics of the diseased retina supported by clinical observations. Our algorithms are trained and tested on images from publicly available, as well as, on our own databases. 

\begin{figure}[!ht]
\centering
\subfigure[]{\includegraphics[keepaspectratio,width=5cm]{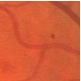}}
\quad\quad
\subfigure[]{\includegraphics[keepaspectratio,width=5cm]{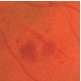}}
\\
\subfigure[]{\includegraphics[keepaspectratio,width=5cm]{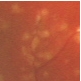}}
\quad\quad
\subfigure[]{\includegraphics[keepaspectratio,width=5cm]{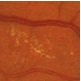}}
\\
\subfigure[]{\includegraphics[keepaspectratio,width=5cm]{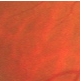}}
\quad\quad
\subfigure[]{\includegraphics[keepaspectratio,width=5cm]{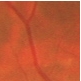}}
\caption{Lesions of the retina; (a) microaneurysms, (b) haemorrhages, (c-d) bright lesions (exudates), (e-f) retinal pigment epihelium.}
\label{fig:ma}
\end{figure}

As a second (pre-filtering) step we extract those candidate subregions of fundus images that are expected to contain specific lesions. This step is economically favourable since the operation of the lesion detection algorithms are computationally the most time-consuming. The most common lesion that algorithms detect on the fundus is the microaneurysm  (see Figure \ref{fig:ma}a), which is an early sign of diabetic retinopathy. A microaneurysm appears as a small red spot on the retina. Microaneurysms may evolve to haemorrhages, which are also red spots, but they differ in size and shape (see Figure \ref{fig:ma}b). The detection of DR related bright lesions (exudates) has a rich literature \cite{Winder} \cite{exudate}, as well. Exudates appear at an advanced stage of diabetic retinopathy (see Figure \ref{fig:ma}c and \ref{fig:ma}d). The retinal pigment epihelium (RPE) is usually caused by age-related macular degeneration. The sign of RPE is the inhomogeneous surface of the retina, as it is shown in Figure \ref{fig:ma}e and \ref{fig:ma}f.

\begin{figure}[!ht]
	\centering
		\includegraphics[keepaspectratio,width=5cm]{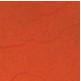}
	\caption{Surface of a normal fundus.}
	\label{fig:normal}
\end{figure}

Our approach aims to find candidate regions containing lesions based on the fact that asides form its anatomical parts, the intensity values of the normal retina surface have small saliences (see Figure \ref{fig:normal}). If there is a connected set of salient values with a given cardinality, we can assume that there is a lesion within the examined region. The goal is to preserve those regions only, which possibly contain lesions.

The rest of the paper is organized as follows. In section \ref{sec:system} we present our pre-screening approach for classifying the images as highly abnormal or not. Section \ref{sec:pref} exhibits how candidate regions that passed the pre-screening are pre-filtered on the fundus images. The datasets and corresponding experimental results are given in section \ref{sec:res}. Finally, some conclusions are drawn in section \ref{sec:concl}. 

\section{\uppercase{Pre-screening -- classifying the input image}}
\label{sec:system}

As first step of our approach, we check whether the fundus represented on the image has so severe abnormalities (e.g. large haemorrhages, retinal detachment) that the patient should be sent directly to clinical expert. In the case of high-loaded automatic systems, skipping these images will enhance the performance, since detailed analyses should not take place. Pre-screening is realized with the application of machine learning algorithms. Next, we summarize the components of pre-screening by consequent stjpg.

\subsection{Pre-processing}
\label{sec:preproc}

\begin{figure}[!ht]

\begin{minipage}[b]{.48\linewidth}
  \centering
 \centerline{\epsfig{figure=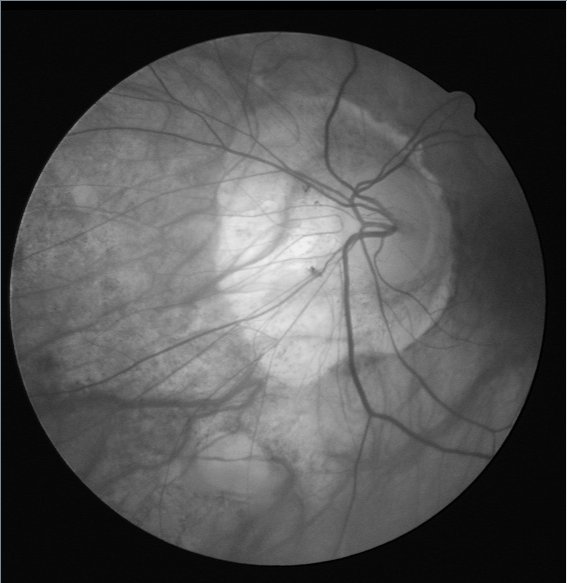,width=5.cm}}
\end{minipage}
\hfill
\begin{minipage}[b]{0.48\linewidth}
  \centering
 \centerline{\epsfig{figure=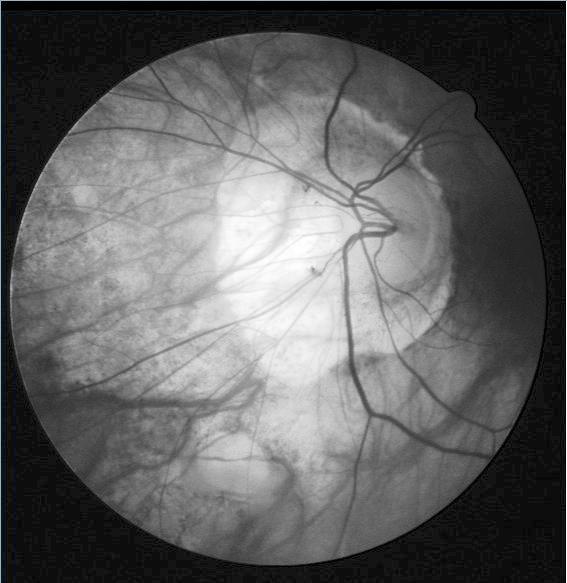,width=5.cm}}
\end{minipage}

\caption{Contrast enhancement of fundus images by adaptive histogram equalization. Original image is taken from the DRIVE database \cite{drive}. (a) grayscale image, (b) image after AHE.}
\label{fig:ahe}

\end{figure}

As a pre-processing step, we convert the input RGB images to grayscale ones as proposed e.g.\ in \cite{extrudate}, 
to get a suitable representation for possible disorders.
Then, we apply adaptive histogram equalization (AHE) as an intensity normalization step proposed in \cite{youssif}, and an output
is shown in Figure \ref{fig:ahe}. Finally, we rescale the images to the size of 90 $\times$ 90 pixels.

\subsection{Feature vectors and classifiers}
\label{sec:fv} 

We also take advantage of the clinical observation 
that fundi with severe diabetic retinopathy often have inhomogeneity caused by retinal 
pigment epithelium (RPE) atrophy, which is the waste of the pigmented cell layer of the 
retina \cite{retina}. Composing feature vectors based on this observation leads to more accurate 
results both in classification and computational performance, as it will be presented in the results section.
To extract these features, we used the following approaches:\\

\begin{itemize}
	\item \textbf{Inhomogeneity:} Let the image be split into disjoint subimages of size $s \times s$, e.g.\ with $s=5$.
Then, for each pixel within a subimage, we compute the sum of intensity differences larger than a given threshold $t$ 
for every subsequent subimage pixels.
If this number is larger than zero, the feature is set to 1, otherwise to 0. See Algorithm \ref{alg:alg1} for the precise formulation. The values of $s$ and $t$ are determined experimentally and these values are constants across the image.
 \item \textbf{Standard deviation:} For each subimage we calculate the standard deviation. This approach is for referential purposes.
 \item \textbf{Combined:} We calculate both the inhomogeneity and the standard deviation feature and combine them.
\end{itemize}
\begin{algorithm}[!ht]

\caption{Inhomogeneity feature calculation.}

\label{alg:alg1}

\begin{algorithmic}

	\STATE{Let $f = 0$.}
	
	\FOR{each subimage $s$ in the image}
	
	\STATE{Let $diff := 0$.}
	
	\FOR{each pixel in $s$}
	
	\STATE{Calculate the differences of the intensities from the first pixel of $s$.}
	
	\IF{the difference is larger than a threshold $t$}
		\STATE{Add the difference to $diff$.}
	\ENDIF
	
		\ENDFOR
		\IF{$diff > 0$}
				
				\STATE{Let $feature\left[f\right] = 1$.}
				
			\ELSE
			
				\STATE{Let $feature\left[f\right] = 0$.}
				
			\ENDIF
		\STATE {Increment $f$.}
		\ENDFOR

%
			
%
%
%
%
%
%
%
%
%
%
%
%
%
%
%
%
%
%

\end{algorithmic}

\end{algorithm}

An example feature vector for a homogeneous and an inhomogeneous image part are given in Table \ref{tab:ex}.

\begin{table}[!ht]
	\centering
		\begin{tabular}{|c|c|c|c|c|c|}
		 \hline
		homogeneous (Figure \ref{fig:normal}) &  4.57, 0.0 & 4.68, 0.0 & 4.34, 0.0 & 3.91, 0.0 & 3.67, 0.0\\
		 \hline
		inhomogeneous (Figure \ref{fig:ma} (e))& 68.55, 1.0 & 71.41, 1.0 &55.30, 0.0 & 65.64, 1.0 & 34.30, 1.0\\ 
		\hline
		\end{tabular}
	\caption{A characteristic example feature vector excerpt for a homogeneous an inhomogeneous retina part. The numbers in each cell are the standard deviation and inhomogeneity values, respectively.}
	\label{tab:ex}
\end{table}

We use the implementations of the Weka \cite{weka} libraries to test the classification of the images using the features above. After investigating several classifiers (SVM, KNN, etc.) we have chosen the Naive Bayes classifier for this task, which uses a simple, but very effective approach to classify in specific cases. As it can be seen in section \ref{sec:res}, this classifier with the proposed feature provided proper results with good computational performance.

It is also an interesting observation that these features (with a proper training set and classifier) are also useful in detecting low quality images. However, this issue is out of the scope of this paper.

\section{\uppercase{Pre-filtering -- extracting regions with lesion candidates}}
\label{sec:pref}

As the second step of our approach, we extract regions with lesions candidate in the images that passed the pre-screening phase. Since these images must undergo detailed image analyses to extract specific lesions later on, this pre-filtering is highly recommended to restrict the input of the corresponding detector algorithms. Now we summarize the stjpg how the candidate regions are extracted.

\subsection{Pre-processing}

\begin{figure}[!ht]
	\centering
		\includegraphics[keepaspectratio,width=\linewidth]{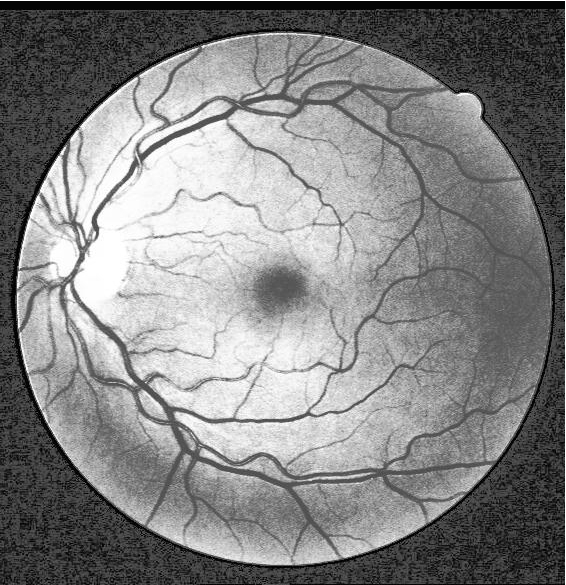}
	\caption{The green plane after histogram equalization.}
	\label{fig:green}
\end{figure}

Similar to the pre-processing stjpg discussed for the pre-screening phase, we use the green plane of the image by following literature recommendations \cite{youssif}. Then, we perform histogram equalization on the image to reduce the vignette effect (see Figure \ref{fig:green})  and calculate the background image by applying a strong median filter of size $A\,\times\,A$ (e.g.\ with $A = 25$). 

\begin{figure}[!ht]
	\centering
		\includegraphics[keepaspectratio,width=\linewidth]{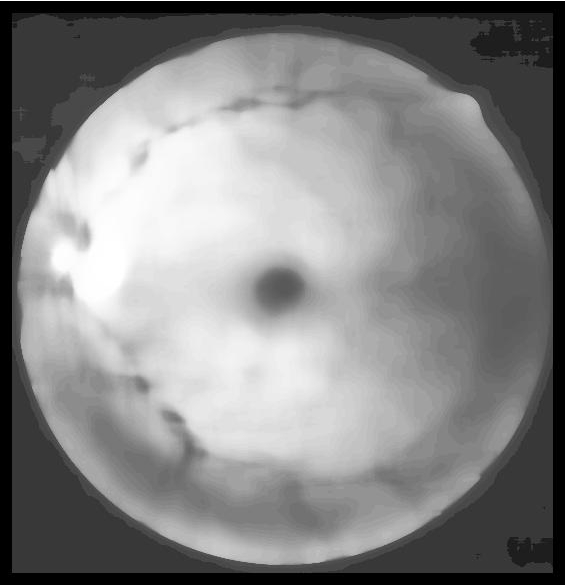}
	\caption{The background image.}
	\label{fig:bg}
\end{figure}

We use the background image shown in Figure \ref{fig:bg}, to perform shade correction by subtracting it from the original image. 

\begin{figure}[!ht]
	\centering
		\includegraphics[keepaspectratio,width=\linewidth]{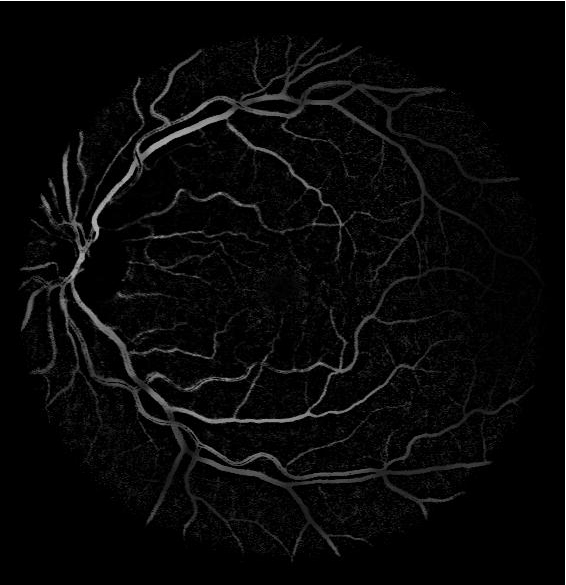}
	\caption{The pre-processed image for candidate region extraction.}
	\label{fig:intrest}
\end{figure}

To suppress noise, we apply a median filter of size $B\,\times\,B$ (e.g.\ with $B = 13$) to the shade corrected image. As the final pre-processing step, we apply unsharp masking to increase the acutance (see Figure \ref{fig:intrest}).  

\subsection{Removal of anatomical parts}

Detecting the anatomical parts of the fundus is an important step before lesion detection. For example, the optic disc appears as the brightest circular patch on the fundus, whose presence may disturb the detection of exudates. Removing the vessel system is also relevant, since a small portion of it appears basically the same as haemorrhages. Besides these two anatomical parts, we also remove the macula, because for certain region sizes, some parts of it can appear as a locally salient object. For these tasks, we use the vessel detector published by Staal et al. \cite{vessel}, the macula detector of Petsatodis \cite{od} and the optic disc detector described in \cite{exudate}. 

\subsection{Statistical analysis of regions}

We split the image into disjoint regions of size $s \times s$ (e.g.\ with $s=75$). For each region, we compute the local mean $\mu$ and the standard deviation $\sigma$ of its intensity values. We label the pixel $P\left(x,\,y\right)$ having intensity $I\left(x,\,y\right)$ as {\em high}, if $I\left(x,\,y\right) -  \mu > \sigma$, while $P\left(x,\,y\right)$ is {\em low}, if $I\left(x,\,y\right) -  \mu  < -\sigma$. Otherwise, {\em P} remains unlabeled. After labelling, we select connected components, which composed of pixels with identical labels and with cardinality at least {\em n}. If a component satisfied these conditions, we consider that as a lesion candidate. We use the areas which possibly contain lesions as input for specific lesion detectors, designed for e.g.\ microaneurysms or exudates.

\section{\uppercase{Results}}
\label{sec:res}

In this section, we present our respective experimental results for the pre-screening and pre-filtering phases of the proposed approach.

\subsection{Results on pre-screening}

Our first experimental dataset consisted of 34 training and 28 test macula-centered images. Both the training and the test sets contain 50-50\% normal and abnormal cases. Ophthalmologists selected and classified these images whether they contain serious disorder or not from three databases: the publicly available DRIVE \cite{drive}, DIARETDB1 \cite{diaretdb1} and the database provided by the Moorfields Eye Hospital, London, UK for our research purposes. Our goal is to find images, where the fundus is abnormal to avoid obviously diseased cases to pass. These images contain sight-threatening disorders and have a grade of R3 in a usual retinopathy grading protocol \cite{protocoll}. Therefore, we label the elements of the test database as images with serious disorder (first class) and images to be processed further (second class). Thus, the second class expected to contain normal or not seriously diseased cases.

For pre-screening, we used a Naive Bayes classifier and trained for the combined features extracted from all regions of the images as disclosed in section \ref{sec:fv}. 
Thus, a $2 \times n$ feature vector is extracted for each image, where $n$ is the number of subimages. With this approach, we have successfully classified all elements of the test dataset. That is, the accuracy in this case is 100\%.  

To make the approach faster, we used backward elimination \cite{hiv} for feature subset selection. That is, we have selected the best 11 regions on each image to extract the features from them for classification. In this case, our approach still provided no false predictions with an elapsed time below milliseconds.

We have also tested our approach on the 1200 images of the Messidor database\footnote{Kindly provided by the Messidor program partners (see http://messidor.crihan.fr)}. This database is dedicated to measure the performance of screening systems by providing grading scores for each image. The grades are from R0 to R3, where R0 represents no retinopathy and r3 is the most serious case of this disease based on the type and the number of the lesions appearing on the images. We have selected the two classes as the follows: abnormal (R3) and images that needs further analysis (R0, R1, R2). Our approach achieved an accuracy of 82\% with 81\% sensitivity and 82\% specificity on this dataset. Most of the error originates from the false classification of R2 cases (52\% of all misclassified images are from this class), while a smaller portion of R1 and R3 images also classified wrongly (25\% and 23\% of all misclassification occurred in these images, respectively).    

\subsection{Results on pre-filtering}
\label{sec:results}

We have tested our approach on those images which have been classified as "to be processed further" by the previous pre-screening phase and the positive samples of the first training set. Thus, this database consisted of 36 images. We have also tested this approach on the 784 images of the Messidor dataset.

The detector missed only 1 fundus image which contained lesions on the first dataset. Our results are summarized in Table \ref{tab:res} in details containing the value of the size parameters $s$, the number of correctly / incorrectly (true / false) identified regions, the number of misclassified images and the percentage of the remaining pixels. An image is considered as misclassified, if it contained at least one lesion, but none was found, or otherwise, if it contains no lesions but at least one was found by the algorithm.

\begin{table}[!ht]
	\centering
		\begin{tabular}{|c|c|c|c|c|}
		 \hline
		 Size ($s$) & True & False & Mis- & Percentage\\
		  & & & classified & \\
		 \hline
		 10 & 24 & 10 & 4 & 0.05\\
		 \hline
		 25 & 26 & 10 & 4 & 0.34\\
		 \hline
		 50 & 25 & 9 & 5 & 1.28\\
		 \hline
		 \textbf{75} & \textbf{27} & \textbf{3} & \textbf{1} & \textbf{2.5}\\
		 \hline
		 100 & 16 & 7 & 5 & 3.47\\
		 \hline
		 200 & 4 & 4 & 5 & 4.82\\
		 \hline
		\end{tabular}
	\caption{Experimental results on pre-filtering. Size - the size of the region window, True - the number of correctly identified regions, False - the number of falsely identified regions, Misclassified - the number of misclassified images, Percentage - the percentage of the number of remaining pixels after this step.}
	\label{tab:res}
\end{table}

\begin{figure}[!ht]
	\centering
		\includegraphics[keepaspectratio,width=\linewidth]{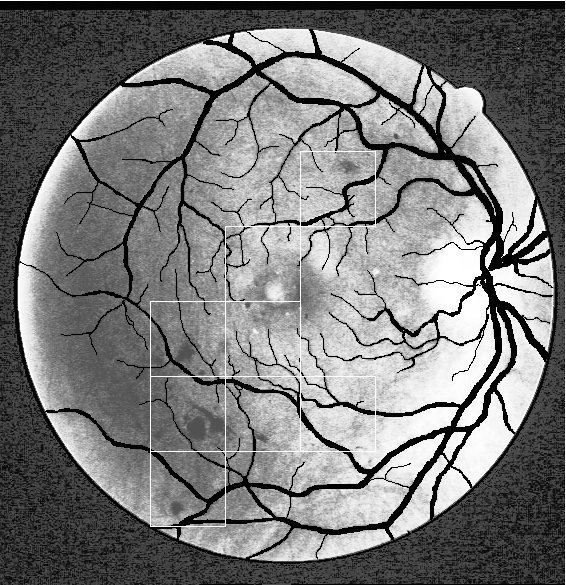}
	\caption{Regions with lesion candidates.}
	\label{fig:result}
\end{figure}

This setup of the parameters was found by empirical tests to obtain highest accuracy. The detector missed only 3.5\% fundus image containing lesions on the first dataset, while on the Messidor dataset this number was 5\%.  
A demonstrative example for selected regions can be seen in Figure \ref{fig:result}. 

With this candidate region detection, we can reduce the total number of pixels of the database to nearly 2.5\% of the original data. In order to demonstrate how its reduction affected consequent detailed image processing analysis, we tested a specific lesion detector. Therefore, the computational time of the state-of-the-art microaneurysm detection algorithm \cite{fleming} reduced by 90\% after this candidate selection.

\section{\uppercase{Conclusion}}
\label{sec:concl}

We presented a complemented automatic decision support approach that can separate fundus images containing serious lesions from the ones that should undergo detailed screening. This step can immediately direct patients with serious lesions to an ophthalmologist without a time-consuming screening procedure. With a use of a Naive Bayes classifier, we were able to classify all the test images correctly. As a secondary pre-filtering step for images passing pre-screening, we have presented an approach which is eligible to detect areas which presumably contain lesions. As a fair trade-off with accuracy, we gained high computational performance with using only small regions to detect the actual lesions within. However, it is also should be noted that with the gain in computational performance some lesions can be missed using this framework.

\section*{\uppercase{Acknowledgment}}

This work was supported in part by the J\'anos Bolyai grant of the Hungarian Academy of Sciences, and by the TECH08-2
project DRSCREEN - Developing a computer based image processing system for diabetic retinopathy screening of the National
Office for Research and Technology of Hungary (contract no.: OM-00194/2008, OM-00195/2008, OM-00196/2008).

\bibliographystyle{elsarticle-num}
\bibliography{refs}

\end{document}